\documentclass[10pt,conference]{IEEEtran}
\IEEEoverridecommandlockouts
\usepackage{cite}
\usepackage{amsmath,amssymb,amsfonts}
\usepackage{algorithmic}
\usepackage{graphicx}
\usepackage{textcomp}
\usepackage{xcolor}
\def\BibTeX{{\rm B\kern-.05em{\sc i\kern-.025em b}\kern-.08em
    T\kern-.1667em\lower.7ex\hbox{E}\kern-.125emX}}
\begin{document}

\title{{\fontsize{24}{29}\selectfont Taming Uncertainty via Automation: Observing, Analyzing, and Optimizing Agentic AI Systems }}
% \title{Conference Paper Title*\\

%\author{\IEEEauthorblockN{Anonymous Authors}}

\author{\IEEEauthorblockN{Dany Moshkovich}
\IEEEauthorblockA{
%\textit{dept. name of organization (of Aff.)} \\
\textit{IBM Research}\\
Haifa, Israel \\
mdany@il.ibm.com}
\and
\IEEEauthorblockN{Sergey Zeltyn}
\IEEEauthorblockA{
%\textit{dept. name of organization (of Aff.)} \\
\textit{IBM Research}\\
Haifa, Israel \\
sergeyz@il.ibm.com}}

\maketitle

\begingroup
\renewcommand\thefootnote{}\footnotetext{International Conference on Automated Software Engineering, ASE 2025.}
\addtocounter{footnote}{-1}
\endgroup

\begin{abstract}

Large Language Models (LLMs) are increasingly deployed within agentic systems—collections of interacting, LLM-powered agents that execute complex, adaptive workflows using memory, tools, and dynamic planning. While enabling powerful new capabilities, these systems also introduce unique forms of uncertainty stemming from probabilistic reasoning, evolving memory states, and fluid execution paths. Traditional software observability and operations practices fall short in addressing these challenges.

This paper presents our vision of AgentOps: a comprehensive framework for observing, analyzing, optimizing, and automating operation of agentic AI systems. We identify distinct needs across four key roles—developers, testers, site reliability engineers (SREs), and business users—each of whom engages with the system at different points in its lifecycle. We present the AgentOps Automation Pipeline, a six-stage process encompassing behavior observation, metric collection, issue detection, root cause analysis, optimized recommendations, and runtime automation. Throughout, we emphasize the critical role of automation in managing uncertainty and enabling self-improving AI systems—not by eliminating uncertainty, but by taming it to ensure safe, adaptive, and effective operation.

\end{abstract}


%%%%%%%%%%%%%%%%%%%%%%%%%%%%%%%%%%%%%%%%%%%%%%%%%%%%%%%%

\section{Introduction}
\label{sec:introduction}

The rapid rise of Large Language Models (LLMs) has ushered in a new generation of agentic AI systems\textemdash where multiple agents collaborate to execute complex, adaptive workflows. These systems execute tasks often defined in natural language, integrate external tools, retain memory, and coordinate dynamic interactions across agents.

Designed for autonomy, agentic systems move beyond fixed instructions to make context-aware decisions using probabilistic reasoning. Their behavior is shaped by LLMs, classical AI planning, and machine learning, often producing divergent outcomes for identical inputs \cite{atil2024non}.

This unpredictability extends to execution paths. Agents can decompose tasks into subtasks, dynamically plan their execution, or delegate them to other agents or tools, which may follow alternative workflows based on internal inference.

Memory introduces additional variation, as agents persist and retrieve information using similarity-based methods\textemdash commonly backed by vector databases\textemdash further amplifying nondeterminism. As they operate, agents accumulate knowledge and develop expertise, making each instance unique and increasingly difficult to update or replace.

The operating environment is equally fluid. Tools may change, vanish, or be introduced at runtime, prompting agents to continuously adapt their usage. In multi-agent settings, coordination emerges from real-time interaction. Like human teams, agents delegate, validate, and reassign roles dynamically\textemdash creating evolving collaboration patterns and feedback loops.

% citation added
This dynamic landscape introduces significant challenges for users—developers, evaluators, SREs, and business stakeholders—across the system lifecycle. Surveys show that only 8\% of organizations use dedicated observability platforms~\cite{precisely, precisely_report}, limiting automation and scalability, and 60\% of users report that current analytics tools do not meet their needs~\cite{moshkovich2025beyond}.

Addressing this uncertainty demands more than traditional operations. It requires a new discipline: \textbf{AgentOps} \cite{agentops_ai,ibm_agentops, dong2024taxonomy}.

\textbf{AgentOps} provides a framework for observing, analyzing, and optimizing intelligent systems that reason and adapt. It treats agents not as static code, but as stateful, evolving entities that must be monitored, guided, and improved\textemdash redefining operations beyond classical ITOps. From instrumentation to feedback loops and self-healing, AgentOps offers the practices needed to operate AI systems safely in enterprise contexts.

The core steps of the AgentOps process include: \textbf{Observing Behavior, Calculating Metrics, Detecting Issues, Identifying Root Causes, Generating Optimized Recommendations}, and \textbf{Automating Operations}\textemdash all supported by a foundation of \textbf{automation}.

AgentOps frameworks are too complex to manage manually\textemdash especially for users lacking machine learning expertise. Even with the right tools, it is difficult to select and apply them correctly. And as systems autonomously adjust prompts, code, or configurations, validating those changes becomes even harder. Automation lowers this barrier by recommending or executing actions to improve system behavior.

This paper makes three contributions. First, it outlines the distinct challenges agentic systems pose for various user roles. Second, it introduces the AgentOps framework taxonomy and maps its components to these roles. Third, it explores automation as a unifying force across roles, analytics pipelines, and self-improving systems.

%%%%%%%%%%%%%%%%%%%%%%%%%%%%%%%%%%%%%%%%%%%%%%%%%%%%%%%%%%%%%%%%%%%%%%%
% 1/3 page
\section{Related Work: Landscape of Agent Analytics}
\label{sec:SOTA}

The agent observability and analytics landscape has split into two categories: GenAI tools \cite{jose2024harnessing} like Phoenix~\cite{arize}, LangFuse~\cite{langfuse}, and LangSmith~\cite{langsmith} targeting developers, and observability platforms like Datadog~\cite{datadog} and IBM Instana~\cite{instana}, which are evolving to support agentic systems for SREs.  

All of these rely on observability and data collection, reinforcing the need for standardized protocols. \textbf{OpenTelemetry (OTel)}\cite{blanco2023practical}, a key standard for logs, traces, and metrics, is being extended to support agent-based workflows. \textbf{OpenLLMetry}\cite{openllmetry} by Traceloop enables observability for frameworks like LangGraph~\cite{langgraph}, CrewAI~\cite{crewai}, and AutoGen~\cite{autogen}. Other notable efforts include \textbf{OpenInference}\cite{openinference} and LangFuse\cite{langfuse}. Still, there are no widely adopted semantic conventions for agentic tracing, and instrumentation for behaviors like planning or reflection remains limited \cite{moshkovich2025otel}.

Beyond observability, analytics tools lack standardized failure taxonomies. While early proposals exist~\cite{deshpande2025trail, pan2025multiagent}, they are not widely adopted. Critical issues—like unintended loops in multi-agent workflows—often go undetected. Graph Neural Networks~\cite{niro2023detecting} offer a promising direction for encoding agentic structure, but failure analysis using them is still nascent.

\textbf{Root cause analysis} remains similarly underdeveloped. Few existing tools effectively capture causal relationships between agent decisions, tool behaviors, and observed failures. Prior work on causal reasoning in process mining \cite{fournier2025business} and causal discovery in agentic systems \cite{fournier2025agentic} offers promising foundations for adaptation in this context.

There is also a gap in \textbf{recommendation systems}. While problems like latency or hallucination may be flagged, actionable suggestions—such as prompt tuning, agent restructuring, or parameter changes—are rare. \textbf{Optimization} support is minimal, with users left to manually adjust trade-offs between cost, latency, and quality.

% SZ - citation addded
Finally, automation remains limited, with only initial efforts to design approaches tailored to agentic systems \cite{sanwouo2025breaking}. Most systems require manual intervention, even for recurring issues, and workflows are rarely refined or adapted automatically.

%%%%%%%%%%%%%%%%%%%%%%%%%%%%%%%%%%%%%%%%%%%%%%%%%%%%%%%%%%%%%%%%%%%%%
\begin{figure*}[t]
  \includegraphics[width=\textwidth]{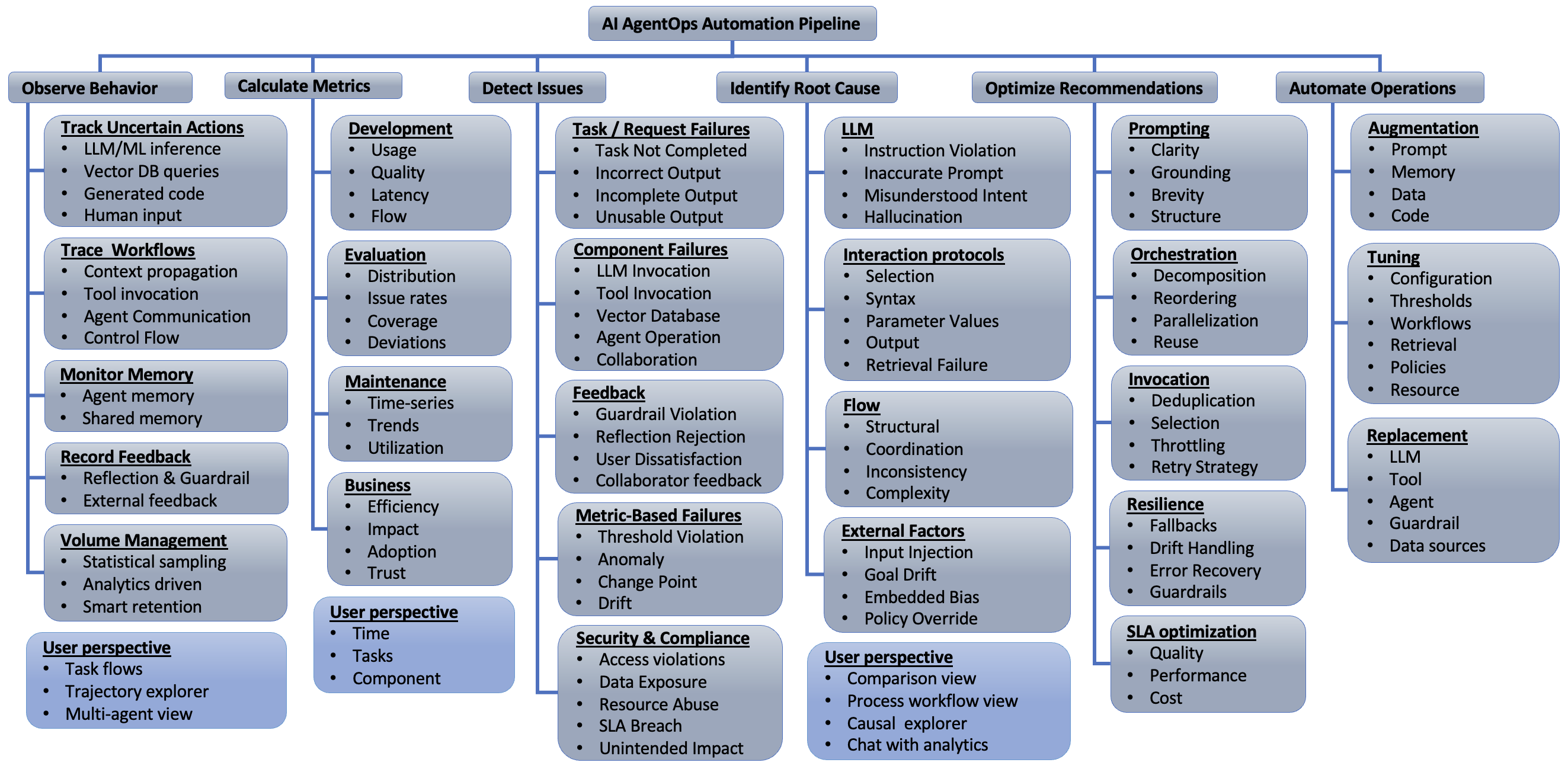}
  \caption{AI AgentOps Automation Pipeline}
  \label{fig:taxonomy}
\end{figure*}

%%%%%%%%%%%%%%%%%%%%%%%%%%%%%%%%%%%%%%%%%%%%%%%%%%%%%%%%%%%%%%%%%%%%%%%%
% 3/4 -pages
\section{Roles and Responsibilities: How Agents Change the Game}

\label{sec:roles}

Agentic systems disrupt boundaries between development, testing, deployment, and operations. Their dynamic, unpredictable nature requires all roles to adapt. We focus on four core roles—developers, testers, operators, and business users—who span the full system lifecycle and, like security, compliance, and product management roles, face distinct challenges as they adapt to agentic reality.

\subsection{Developers}

In traditional software development, developers design and implement code, then debug it using breakpoints and by testing a few fixed execution paths. In agentic development, however, this approach is no longer sufficient\textemdash running the system once, even with the same initial conditions, may yield different outcomes due to inherent stochasticity and dynamic reasoning.

Developers must tune numerous LLM parameters (e.g., temperature, context window) and configure integrated tools and APIs\textemdash often under tight coupling and evolving constraints. Designing and orchestrating dynamic behavior in agentic systems under these conditions is particularly challenging.

Understanding system behavior requires instrumentation to capture structured trace data. In parallel, prompt engineering introduces a vast, unstructured design space that demands iterative experimentation. Dynamic code generation further complicates this, as execution flows can shift at runtime, requiring greater adaptability.

\subsection{Testers} 

Testers are responsible for comprehensive validation, traditionally measured by \textbf{code} and \textbf{requirements coverage}. In agentic systems, however, non-determinism complicates this task\textemdash executing all code paths and requirements does not guarantee coverage of all relevant behaviors.

Outcomes are often non-binary, with success on a \textbf{continuous spectrum} and acceptable thresholds varying by application criticality. Testing must shift from final outputs to \textbf{intermediate states} and \textbf{decision points}—such as tool choices or routing logic—to fully understand behavior.

Finally, evaluation should not end at deployment. As agentic systems evolve with usage, context, and learning, post-deployment testing and monitoring are essential for sustained reliability, trustworthiness, and performance.

\subsection{Site Reliability Engineers}

A Site Reliability Engineer (SRE) bridges software engineering and operations, focusing on building systems that are both reliable and efficient. Traditionally, SREs monitor performance metrics and respond to failures or outages.

In agentic systems, the focus shifts to \textbf{proactive trend analysis}. Like physicians tracking vital signs, SREs must monitor numeric and semantic indicators\textemdash such as latency, cost, tool usage, and human input\textemdash to detect early signs of systemic issues. Waiting for a complete system failure is akin to diagnosing a heart attack after it occurs\textemdash by that point, intervention may be too late.

When a drift, anomaly, or inefficiency is detected, SREs must perform \textbf{root cause analysis} and identify actionable mitigation strategies. Some issues can be addressed automatically\textemdash through dynamic reconfiguration, scaling, prompt changes, or tool replacement\textemdash while others require developer intervention. This closes the loop between observability and ongoing system improvement.

\subsection{Business Users}

Unlike SREs, business users focus on \textbf{business-centric metrics} such as revenue, cost, ROI, and customer satisfaction—especially the trade-off between cost, latency, and quality. These metrics must be monitored, linked to business outcomes, and analyzed for anomalies and root causes.

Beyond issue detection, business users explore new business opportunities through what-if analyses and A/B testing, to support strategic decision-making.

%%%%%%%%%%%%%%%%%%%%%%%%%%%%%%%%%%%%%%%%%%%%%%%%%%%%%%%
% 1.5 pages
% 
\section{AI AgentOps Automation Pipeline}
\label{sec:agentops}

To address emerging challenges faced by developers, testers, operators, and business users, we introduce the AI AgentOps Automation Pipeline—a six-stage process spanning from raw behavior capture to self-healing mechanisms. Each stage targets a core operational need and supports both automated and user-driven methods. The accompanying taxonomy and user perspectives (Fig.~\ref{fig:taxonomy}) illustrate these concepts; below, we explain their rationale and interconnections.

\subsection{Observe Behavior}

Observing an agentic system goes beyond classical tracing—it requires capturing how decisions and execution flows emerge dynamically, including those driven by code generated by the agent at runtime. This involves automatic instrumentation to trace probabilistic behaviors such as LLM inference, tool usage, vector database queries, and human input.

Understanding workflows requires tracking context propagation, tool invocations, and inter-agent communication to reconstruct decision paths—especially when control flow is composed on the fly.

Observability must also capture feedback loops, including internal reflection, guardrails, and user input, that influence behavior. To manage trace volume, automatic volume management techniques such as statistical sampling, anomaly-triggered escalation, and smart retention help maintain visibility while limiting overhead.

Finally, observability should aid interpretation through task flow visualizations, execution trajectories, and multi-agent views that reveal how behavior evolves at runtime.

\subsection{Collect Metrics}

While observability surfaces raw signals, metrics automatically transform them into structured insights. During development, we track usage metrics like tool call frequency and memory access rate, quality indicators such as task success and output completeness, as well as latency and flow characteristics including task volume, branching complexity, and reasoning depth.

Evaluation emphasizes behavior distribution across runs, failure and degradation rates, functional and edge case coverage, and deviations from golden traces or ground truth.

In maintenance, time-series metrics reveal regressions, drift trends, and system utilization—supporting proactive scaling and adaptation.

For business stakeholders, metrics highlight efficiency through cost and latency, impact via ROI and productivity, adoption through usage penetration, and trust via positive feedback and compliance adherence.

User-scoped metrics enable slicing by time intervals, task categories, and system components, aligning insights with specific user needs.

\subsection{Detect Issues}

The automation layer analyzes data and metrics to detect issues—including both outright failures and subtle degradations. It categorizes them by type and scope, assigns severity, correlates related events, and triggers smart alerts.

Task failures cover incomplete tasks, incorrect or partial outputs, and syntactically valid but unusable responses.

Component issues may not block task completion but reveal deeper, often systemic problems—such as LLM timeouts, low-confidence outputs, tool or vector DB errors, and agent miscoordination.

Failures in handling feedback—such as violations of guardrails, rejection of reflective feedback, user dissatisfaction, or collaboration failures—are another critical class. These break the adaptive loop that makes agents responsive and intelligent.

Metric-based failures manifest when monitored values breach pre-defined thresholds, take on anomalous values, hit change points indicating discrete shifts in behavior, or exhibit pattern shifts that signal evolving system dynamics.

Security and compliance issues cannot be ignored. These include breaches of access control policies, exposure of sensitive data, misuse of resources, SLA violations, and unintended side effects that compromise operational or ethical integrity.

\subsection{Identify Root Cause}

Root cause analysis (RCA) automatically bridges the gap between symptoms and solutions. A common root cause category involves LLM-related issues—such as instruction violations, ambiguous or inaccurate prompts, misunderstood intent, or hallucinations. Even subtle prompt flaws can lead to significant behavioral differences.

Problems in interaction protocols—both between agents and in tool invocation—often stem from improper tool or parameter selection, syntax errors, and invalid configurations. These missteps can lead to breakdowns in execution, particularly when the necessary context is missing or retrieval fails.

Flow and coordination failures reflect deeper behavioral problems: misaligned task decomposition, inconsistent coordination between agents, structural gaps, or overly complex multi-step goals.

External factors also play a role. Input injection can derail planning, goal drift shifts system behavior away from intended outcomes, embedded bias may skew decisions, and policy overrides may suppress needed checks.

To support investigation, users benefit from comparison views between healthy and failing traces, end-to-end process workflow reconstructions, causal path explorers that surface potential dependencies, and analytics chat interfaces that allow direct queries like "Why did this fail?"

\subsection{Optimize Recommendations}

Once root causes are known, optimization focuses on making targeted improvements. Prompting issues are often addressed first—by clarifying ambiguous phrasing, grounding responses in relevant context, tightening verbosity, or applying clearer structural patterns.

Workflow-level enhancements involve refining task decomposition, reordering steps for efficiency, enabling parallelization, and reusing results when appropriate. On the invocation side, removing redundant calls, selecting better tools, applying throttling, and using smarter retry logic can stabilize execution.

To improve resilience, systems should incorporate fallback options, detect and manage behavioral drift, recover gracefully from errors, and enforce guardrails. All these adjustments must also consider SLA tradeoffs, balancing quality, performance, and cost in line with user priorities.

\subsection{Automate Operations}

Automation closes the loop by enacting improvements automatically when confidence is high. This includes augmentation of prompts, runtime data, and tool instructions to adjust agent behavior on the fly, as well as tuning of configurations, thresholds, retrieval logic, or timeouts to maintain optimal performance.

When deeper issues persist, AgentOps may switch LLMs, replace tools, modify workflows, update guardrails, or reset faulty components—all without requiring code changes or redeployment.

For example, AgentOps \textbf{observes behavior} by noticing that an SRE Agent \cite{jha2025itbench} inconsistently uses a new diagnostic tool, based on traces of its \textit{workflow execution} and \textit{tool invocations}. AgentOps \textbf{collects metrics}, monitoring \textit{issue rates} and identifying frequent \textit{incomplete outputs} for a specific \textit{task}. Next, AgentOps \textbf{detects issues} by spotting a \textit{drift} in this failure pattern, indicating rising \textit{task failures}. AgentOps \textbf{identifies the root cause} as \textit{inaccurate prompt} instructions and \textit{tool misuse}. It then \textbf{optimizes recommendations} by suggesting a clearer \textit{prompt}. Finally, AgentOps \textbf{automates operation} by \textit{augmenting} the Agent’s \textit{prompt} and revalidating the fix through continuous monitoring—automatically optimizing behavior without changing code. Like a manager guiding an employee based on performance feedback, AgentOps enables agentic systems to self-correct and improve in real time.

\section{Discussion: Taming, not Eliminating, Uncertainty}

\label{sec: discussion}
 Uncertainty is intrinsic to intelligence. Just as we accept ambiguity in human reasoning, we must also recognize it in intelligent software systems. But recognition does not imply surrender. While agentic systems will inevitably exhibit behavioral uncertainty, the goal is to tame it—minimizing the frequency and severity of undesirable or strongly suboptimal outcomes.

Promising directions for taming uncertainty through automation include:

\begin{itemize}
\item \textbf{Standardization.} Our taxonomy provides a foundation for AgentOps instrumentation, evaluation, and automation, aligned with emerging standards such as OpenTelemetry, MCP~\cite{hou2025mcp}, and the UIM Protocol~\cite{uim} that promote interoperability across agentic systems.
\item \textbf{Graph-based, alphanumeric analytics.} Agentic systems produce structured, graph-shaped data with semantic richness. New methods must encode and apply this data for issue detection and root cause analysis.
\item \textbf{Self-healing and adaptive execution.} Automated mechanisms should enable systems to respond to problems in real time—rerouting tasks, adjusting LLM parameters, or altering execution plans—reducing the impact of suboptimal behavior without always requiring human intervention.

\end{itemize}

\newpage

\bibliographystyle{IEEEtran}
\bibliography{agent_analytics}

@article{atil2024non,
  title={Non-determinism of "deterministic" {LLM} settings},
  author={Atil, Berk and Aykent, Sarp and Chittams, Alexa and Fu, Lisheng and Passonneau, Rebecca J and Radcliffe, Evan and Rajagopal, Guru Rajan and Sloan, Adam and Tudrej, Tomasz and Ture, Ferhan and others},
  journal={arXiv preprint arXiv:2408.04667},
  year={2024}
}

@book{blanco2023practical,
  title={Practical OpenTelemetry: Adopting Open Observability Standards Across Your Organization},
  author={Blanco, Daniel Gomez},
  year={2023},
  publisher={Springer},
  address={Heidelberg, Germany}
}

@article{deshpande2025trail,
  title={TRAIL: Trace Reasoning and Agentic Issue Localization},
  author={Deshpande, Darshan and Gangal, Varun and  Mehta, Hersh and Krishnan, Jitin and Kannappan, Anand and Qia, Rebecca},
  journal={arXiv preprint arXiv:2505.08638},
  year={2025},
}

@article{dong2024taxonomy,
  title={A Taxonomy of {AgentOps} for Enabling Observability of Foundation Model based Agents},
  author={Dong, Liming and Lu, Qinghua and Zhu, Liming},
  journal={arXiv preprint arXiv:2411.05285},
  year={2024},
}

@article{fournier2025agentic,
  title={Agentic {AI} Process Observability: Discovering Behavioral Variability},
  author={Fournier, Fabiana and Limonad, Lior and David, Yuval},
  journal={arXiv preprint arXiv:2505.20127},
  year={2025}
}

@article{fournier2025business,
  title={The WHY in Business Processes: Discovery of Causal Execution Dependencies},
  author={Fournier, Fabiana and Limonad, Lior and Skarbovsky, Inna and David, Yuval},
  journal={KI-K{\"u}nstliche Intelligenz},
  pages={1--23},
  year={2025},
  publisher={Springer}
}

@article{hou2025mcp,
  title={Model context protocol ({MCP}): Landscape, security threats, and future research directions},
  author={Hou, Xinyi and Zhao, Yanjie and Wang, Shenao and Wang, Haoyu},
  journal={arXiv preprint arXiv:2503.23278},
  year={2025}
}

@article{jha2025itbench,
  title={Itbench: Evaluating {AI} agents across diverse real-world it automation tasks},
  author={Jha, Saurabh and Arora, Rohan and Watanabe, Yuji and Yanagawa, Takumi and Chen, Yinfang and Clark, Jackson and Bhavya, Bhavya and Verma, Mudit and Kumar, Harshit and Kitahara, Hirokuni and others},
  journal={arXiv preprint arXiv:2502.05352},
  year={2025}
}

@article{jose2024harnessing,
  title={Harnessing Large Language Models ({LLMs}) Optimizing Performance, Monitoring, and Compliance},
  author={Jose, Edwin and Prabhakaran, Prasad},
  journal={Authorea Preprints},
  year={2024},
  publisher={Authorea},
}

@article{moshkovich2025beyond,
  title={Beyond black-box benchmarking: Observability, analytics, and optimization of agentic systems},
  author={Moshkovich, Dany and Mulian, Hadar and Zeltyn, Sergey and Eder, Natti and Skarbovsky, Inna and Abitbol, Roy},
  journal={arXiv preprint arXiv:2503.06745},
  year={2025}
}

@inproceedings{niro2023detecting,
  title={Detecting anomalous events in object-centric business processes via graph neural networks},
  author={Niro, Alessandro and Werner, Michael},
  booktitle={International Conference on Process Mining},
  pages={179--190},
  year={2023},
  organization={Springer}
}

@inproceedings{pan2025multiagent,
  title={Why do multiagent systems fail?},
  author={Pan, Melissa Z and Cemri, Mert and Agrawal, Lakshya A and Yang, Shuyi and Chopra, Bhavya and Tiwari, Rishabh and Keutzer, Kurt and Parameswaran, Aditya and Ramchandran, Kannan and Klein, Dan and others},
  booktitle={ICLR 2025 Workshop on Building Trust in Language Models and Applications},
  year={2025}
}

@inproceedings{sanwouo2025breaking,
  title={Breaking the Loop: {AWARE} is the new {MAPE-K}},
  author={Sanwouo, Brell Peclard and Temple, Paul and Quinton, Cl{\'e}ment},
  booktitle={FSE'25-International Conference on the Foundations of Software Engineering},
  year={2025}
}

@online{agentops_ai,
  author={{AgentOps.ai}},
  title={Trace, Debug, \& Deploy Reliable {AI} Agents},
  year={2025},
  url={https://www.agentops.ai/},
  urldate={2025-06-29},
  organization={AgentOps.ai}
}

@online{arize,
  author={{Phoenix}},
  title={Phoenix},
  year={2025},
  url={https://phoenix.arize.com/},
  urldate={2025-06-29},
  organization={Arize AI}
}

@online{autogen,
  author={{AutoGen}},
  title={AutoGen},
  year={2025},
  url={https://microsoft.github.io/autogen/stable/},
  urldate={2025-07-01},
  organization={Microsoft}
}

@online{crewai,
  author={{CrewAI}},
  title={{CrewAI}},
  year={2025},
  url={https://www.crewai.com/},
  urldate={2025-06-29},
  organization={CrewAI}
}

@online{datadog,
  author={{Datadog}},
  title={Datadog},
  year={2025},
  url={https://www.datadoghq.com/},
  urldate={2025-06-29},
  organization={Datadog}
}

@online{ibm_agentops,
  author={{Murphy, Mike}},
  title={How to know if your {AI} agents are working as intended},
  year={2025},
  url={https://research.ibm.com/blog/ibm-agentops-ai-agents-observability},
  urldate={2025-07-10},
  organization={IBM Research}}

@online{instana,
  author={{IBM Instana}},
  title={{IBM} {Instana}},
  year={2025},
  url={https://www.ibm.com/products/instana},
  urldate={2025-07-10},
  organization={IBM}
}

@online{langfuse,
  author={{LangFuse}},
  title={LangFuse},
  year={2025},
  url={https://langfuse.com/},
  urldate={2025-06-29},
  organization={LangFuse}
}

@online{langgraph,
  author={{LangGraph}},
  title={LangGraph},
  year={2025},
  url={https://www.langchain.com/langgraph},
  urldate={2025-06-29},
  organization={LangChain}
}

@online{langsmith,
  author={{LangSmith}},
  title={LangSmith},
  year={2025},
  url={https://smith.langchain.com/},
  urldate={2025-02-12},
  organization={LangChain}
}

@online{openinference,
  author={{OpenInference}},  
  title={{OpenInference}},
  year={2025},
  url={https://arize.com/docs/ax/learn/tracing-concepts/what-is-openinference},
  urldate={2025-06-29},
  organization={Arize AI}
}

@online{openllmetry,
  author={{OpenLLMetry}},  
  title={OpenLLMetry},
  year={2025},
  url={https://www.traceloop.com/openllmetry},
  urldate={2025-06-29},
  organization={Traceloop}
}

@online{precisely,
  author={{Skeen, Julie}},  
  title={Mastering {AI} Data Observability: Top Trends and Best Practices for Data Leaders},
  year={2025},
  url={https://www.precisely.com/blog/data-quality/mastering-ai-data-observability-top-trends-and-best-practices-for-data-leaders},
  urldate={2025-08-31},
  organization={Precisely}
}

@online{precisely_report,
  author={{Precisely}},  
  title={{BARC} Research Study: Observability for {AI} Innovation},
  year={2025},
  url={https://www.precisely.com/resource-center/analystreports/barc-research-study-observability-for-ai-innovation},
  urldate={2025-08-31},
  organization={Precisely}
}

@online{uim,
  author={{Synapti.ai}},  
  title={Unified Intent Mediator Protocol},
  year={2025},
  url={https://www.uimprotocol.com/},
  urldate={2025-07-07},
  organization={Synapti.ai}
}

@online{moshkovich2025otel,
  author={Moshkovich, Dany},
  year={2025},
  title={Semantic Conventions for Generative {AI} Agentic Systems},
  url={https://github.com/open-telemetry/semantic-conventions/issues/2664},
  urldate={2025-09-02}
}
\vspace{12pt}

\end{document}